# A Context-Enhanced De-identification System

Kahyun Lee, M.S.
  George Mason University, Fairfax, VA, USA, klee70@gmu.edu

Mehmet Kayaalp, M.D., Ph.D.
  U.S. National Library of Medicine, Bethesda, MD, USA, mkayaalp@mail.nih.gov

Sam Henry, Ph.D.
  George Mason University, Fairfax, VA, USA, shenry20@gmu.edu

Özlem Uzuner, Ph.D.
  George Mason University, Fairfax, VA, USA, ouzuner@gmu.edu

**ABSTRACT**

Many modern entity recognition systems, including the current state-of-the-art de-identification systems, are based on bidirectional long short-term memory (biLSTM) units augmented by a conditional random field (CRF) sequence optimizer. These systems process the input sentence by sentence. This approach prevents the systems from capturing dependencies over sentence boundaries and makes accurate sentence boundary detection a prerequisite. Since sentence boundary detection can be problematic especially in clinical reports, where dependencies and co-references across sentence boundaries are abundant, these systems have clear limitations. In this study, we built a new system on the framework of one of the current state-of-the-art de-identification systems, NeuroNER, to overcome these limitations. This new system incorporates context embeddings through forward and backward $n$-grams without using sentence boundaries. Our context-enhanced de-identification (CEDI) system captures dependencies over sentence boundaries and bypasses the sentence boundary detection problem altogether. We enhanced this system with deep affix features and an attention mechanism to capture the pertinent parts of the input. The CEDI system outperforms NeuroNER on the 2006 i2b2 de-identification challenge dataset, the 2014 i2b2 shared task de-identification dataset, and the 2016 CEGS N-GRID de-identification dataset ($p < 0.01$). All datasets comprise narrative clinical reports in English but contain different note types varying from discharge summaries to psychiatric notes. Enhancing CEDI with deep affix features and the attention mechanism further increased performance.

**CCS CONCEPTS**

• Computing methodologies~Artificial intelligence~Natural language processing~Information extraction • Applied computing~Life and medical sciences~Health informatics

**KEYWORDS**

de-identification, HIPAA, entity recognition, information extraction, natural language processing

## 1 Introduction and Related Work

De-identification refers to detecting and redacting personally identifiable information (PII). The Safe Harbor method of the Health Insurance Portability and Accountability Act (HIPAA) of the United States lists 18 types of PII. As PII in health records links health information to the patient, patient's relatives or patient's employers, the health record becomes protected health information. PII must be removed from health records so that health information can be used for scientific purposes without breaching the patient's privacy. This is not necessary if the patient's consent is given and an approval from the institutional review board is granted. However, obtaining consent from all patients mentioned in a dataset containing years of data is often impracticable. In such cases, de-identification is an essential privacy protection tool, prerequisite for medical research.

Algorithmic text de-identification is an area of natural language processing (NLP) and artificial intelligence (AI). More specifically, de-identification is a named entity recognition (NER) problem [20,26]. The goal of NER is to identify the text spans of entities and determine the type of entity mentioned in those spans. Entity recognition systems have been used for a wide spectrum of tasks ranging from question answering [27,42] to document summarization [5,29]. In the medical domain, example entity recognition tasks include medication information extraction [40,44] and adverse drug event detection [16,20].

Like many AI and NLP methods, text de-identification can be performed either symbolically, using linguistic and domain knowledge for inference, or subsymbolically, relying heavily on the data without using domain knowledge explicitly. Symbolic NLP systems [11,17,39,41] are knowledge-based systems that draw inferences through domain and linguistic knowledge encoded in the algorithm or represented explicitly in rules and variables. Many de-identification systems that learn production rules [4,6] or decision trees [32] are of this nature.

Most machine learning systems [9,13,22,23,53] used for text de-identification are data-driven systems. These systems use mostly surface structures (i.e., lexical items such as words, characters, and their morphological features) to discover patterns in the text without explicitly representing the patterns in their algorithms. Although designing such machine learning systems (e.g., deep neural networks) does not require domain specific knowledge, these systems require large amounts of human-annotated training data for the specific task [36,37,43]. Given a sufficiently large and representative training dataset, the trained model may achieve a higher performance than a human expert [35].

In this study, our focus is on recurrent neural networks (RNNs) with long short-term memory (LSTM) nodes [14], which were shown to be very successful on clinical text de-identification [7,24,54]. RNNs handle data sequentially, predicting the output from the input based on the past input sequences.

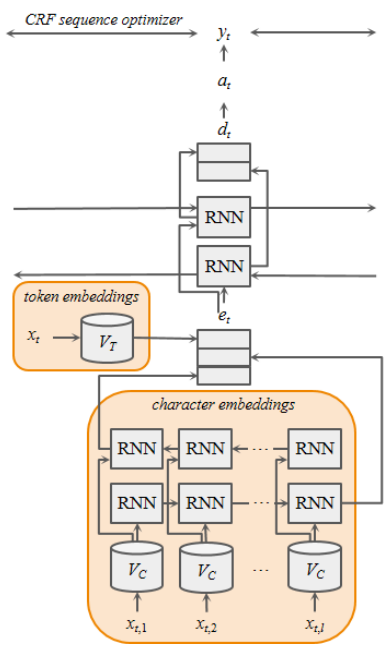

Figure 1: The architecture of biLSTM-CRF-based NER systems at token $x_t$. Token $x_t$ consists of $l$ characters. $V_C$ and $V_T$ are the mappings from character-to-character embeddings and from token-to-token embeddings, respectively. BiLSTM output over character embeddings is concatenated with pre-trained token embeddings $e_t$. The concatenated embedding is fed into another biLSTM layer to produce $d_t$. The probability vector $a_t$ is produced using $d_t$ and the predicted label $y_t$ is adjusted with CRF sequence optimizer.

Lample et al. [18] suggested an NER system based on a bidirectional LSTM (biLSTM) using word and character embeddings as features. Their system also employs a conditional random field (CRF) model for sequence optimization. As shown in Figure 1, their system concatenates biLSTM outputs over character embeddings with token embeddings and feeds the concatenated embeddings into the token-level biLSTM layer sentence by sentence. The CRF sequence optimizer maximizes the likelihood of the label sequence of the sentence, based on these token-level biLSTM outputs. This is the fundamental model of many modern NER systems [12,15,50], including the current state-of-the-art de-identification systems such as [7,24,54]. These de-identification systems take health records as input, split each health record into sentences, tokens, and characters subsequently, and map each character and token to the corresponding embeddings. The embeddings then are processed with biLSTM and CFR sequentially. The systems evaluate the likelihood of token being PII sentence by sentence.

All de-identification systems based on biLSTM-CRF share some common problems: (1) As Finkel and Manning [10] pointed out, the performance of an NLP system is significantly affected by the quality of lower level task such as sentence boundary detection which is often performed as part of pre-processing the input. BiLSTM-CRF-based de-identification systems make a strong assumption on the order of tokens in the input sentence. If the first $m$ tokens of the input are not the first $m$ words of the sentence (i.e., if the assumption does not hold), the system performance would be degraded. However, partitioning text into sentences is an imperfect process, especially for narrative clinical reports with informal, highly abbreviated content, containing an abundance of acronyms and punctuation errors. (2) Even if the text is partitioned into sentences perfectly, these systems cannot capture dependencies outside of sentence boundaries because they treat sentences independently [25]. In other words, these de-identification systems can miss PIIs due to the lack of context provided in the adjacent sentences. (3) Since lengths of sentences are heterogeneous and each token input is received via a fixed node depending on the location of the token in the sentence, input layer representations of similar but non-identical sentences are inconsistent and learning collocational relations between equidistant tokens is harder. (4) For de-identification systems that use pre-trained token embeddings, the above problems are exacerbated by ambiguous or out-of-vocabulary (OoV) tokens, which are common.

## 2  Methods and Materials

In this study, we propose a novel approach for de-identification: a context-enhanced de-identification (CEDI) system, which bypasses the problem of sentence boundary detection altogether. CEDI captures context information via a moving window of $n$-grams rather than in discrete sentences for which token input locations at the input layer are fixed. Although the moving window of $n$-grams has been used previously for clinical relation extraction [21], this is the first time it is used to overcome the limitations such as the sentence boundary detection, dependencies outside of sentence boundaries, heterogeneous length of sentences, and OoV tokens in de-identification systems. We built CEDI on the framework of NeuroNER, which uses only character and token embeddings as features. We add, as an additional set of features, automatically derived context embeddings to the base system to capture dependencies that fall outside of sentence boundaries. These context embeddings are automatically generated using forward and backward $n$-grams. Unlike other context embeddings such as Embeddings from Language Model (ELMo) [31] that require pre-training outside of the NER system and usually in a different domain, CEDI derives the $n$-gram context embeddings internally with no prerequisite training, creating consistent input layer representations and generating embeddings for ambiguous and OoV tokens as well. We also introduce deep affix features [52] and an attention mechanism [2] to CEDI for further improvement. These have been used in other NER tasks [33,49,52,55], but not for text de-identification.

### 2.1 $n$-gram context embeddings

Figure 2 shows how CEDI produces context embeddings using $n$-grams. First, for each token $x_t$, $n$ preceding tokens $(x_{t-n}, \cdots, x_{t-1})$ are extracted. Then the randomly initiated token embeddings for the $n$-gram of tokens $(V_{x_{t-n}}, \cdots, V_{x_{t-1}})$ followed by the token embedding $V_{x_t}$ are loaded. The loaded token embeddings are read sequentially by the corresponding RNN units. The output from the last RNN unit, of which the corresponding token embedding was trained for token $x_t$, is the forward context embedding for the token. The same process is repeated with the backward $n$-gram $(x_{t+n}, \cdots, x_{t+1})$ followed by $x_t$ to derive the backward context embedding. The final context embedding to be fed into the next layer is produced by concatenating the forward and backward context embeddings. In the following layer, the context embeddings can be used alone or concatenated with other embeddings such as character embeddings, pre-trained token embeddings and even other context embeddings depending on the task.

CEDI has several advantages over the prevailing approach. First, context embeddings using $n$-grams do not require sentence boundaries; thus, it is not affected by the problem of erroneous sentence boundary detection.

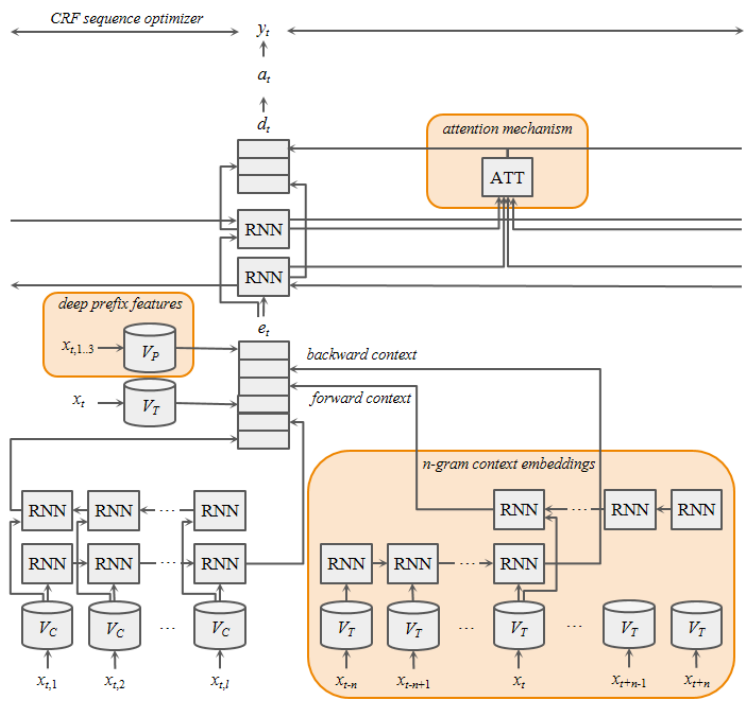

**Figure 2**: A structure of our proposed system at token $x_t$. The $n$-gram context embeddings, deep prefix feature and attention mechanism are integrated with conventional character embeddings and token embeddings. Our context embeddings with $n$-grams is shown in the bottom-right shaded area. Unlike the conventional approach that loads only one pre-trained word embedding per token, our approach loads $2n + 1$ embeddings per token and processes via RNN units to derive $n$-gram context embeddings. The feature embedding $e_t$ is the concatenation of character-, token-, and context-embeddings along with deep prefix features of token $x_t$. The output of the attention mechanism is merged with the biLSTM outputs over the concatenated embedding and then merged into vector $d_t$, which is used to calculate $a_t$, the probability vector over labels.

Second, it captures the context by receiving the input through a $2n + 1$ words wide window rather than through a varying size window delimited by two sentence boundaries. The fixed size context window provides consistency in input representation between training and testing.

Third, since the input sequence moves over the input layer, each token is not associated with a single input node; rather, it traverses through all nodes, making the system learn the relation between the tokens with respect to their relative distances.

Fourth, CEDI can detect dependencies up to $n$ words from either side of the current token word; thus, the size of the context window is adjustable according to the need. An optimal window size can be learned through cross-validation during parameter tuning or can be adjusted a priori according to the nature of the text. For example, the requisite width of the context window can be set wider for psychiatric reports than the window width for nursing notes since sentences in psychiatric reports are much longer and dependencies between tokens often require a wider context for entity recognition.

Fifth, CEDI's approach allows connections between the current token and the information mentioned in the previous or next sentence. For example, the co-reference resolution of the subject of the sentence (e.g., "It happened…") cannot be resolved using the prevailing approach but it is quite possible with CEDI.

Sixth, pre-trained embeddings suffer from OoV terms (terms that do not exist in the pre-trained embeddings) as well as from homonymy (words such as "bat" with identical lexical form but with different meanings); however, CEDI alleviates both problems via contextualization. In pre-trained word embedding approaches, embeddings of two homonymous words are identical even though they have different meanings. Since CEDI does not take the word as a singleton isolated from everything else, it produces embedding representations for each input word based on its surrounding words; thus, homonymous words are associated with distinct embeddings.

In case of OoV terms, the prevailing practice is to assign them non-informative random embeddings for token-level embeddings and use character embeddings in addition to token-level embeddings. While original NeuroNER mitigated problems from OoV on i2b2 2014 dataset using character embeddings, the results of an earlier study on i2b2 2016 dataset [19] indicated that this particular strategy was insufficient for overcoming the OoV problems on this new dataset. CEDI associates every OoV word with their corresponding context embedding, providing meaning to the OoV word using the information of the neighboring words. Thus, the resulting representation of the word is not arbitrary but a function of the surrounding words.

Another context embeddings model, ELMo, has been suggested by Peters et al. [31], and in our previous work [19], we showed how ELMo can improve the de-identification performance of NeuroNER. ELMo is based on bidirectional language models. While ELMo is trained using a separate system (usually in a different and more general context), CEDI derives $n$-gram context embeddings internally; thus, there would be no context shifting with CEDI.

ELMo has several limitations compared to CEDI. To utilize a pre-trained ELMo model, the user needs to process the entire dataset and extract ELMo embeddings before starting NER training. The extraction process took over 6 hours for us for 2014 i2b2 shared task. Because all individual tokens have different context embeddings depending on the sentence, the storage and memory requirements for ELMo can also be overwhelming. For example, in the 2014 i2b2 shared task, the entire extraction process required 30.3 GB of RAM and the extracted ELMo embeddings consumed 4.1 GB of storage—almost 500 times as large as the entire 2014 i2b2 shared task dataset, which was only 8.1 MB. Moreover, the extracted ELMo embeddings cannot be applicable to other datasets. In other words, the user needs to reiterate this extraction process for each dataset separately. In contrast, CEDI derives $n$-gram context embeddings internally, saving the user from ELMo's prerequisites, a lengthy extraction process and extra storage for embeddings.

## 2.2 Deep affix features

Proposed by Yadav et al. [52], deep affix features posit that the first few characters (the approximate prefix) and the last few characters (the approximate suffix) are semantically informative. This notion is particularly relevant to the medical domain. For example, new or uncommon drug names can be OoV words, which are difficult for neural network-based NER systems to capture because they cannot be included in the pre-trained token embeddings. Their labels are therefore predicted from random embeddings. However, if the drug name ends with "-parin" and shares the same suffix embeddings with existing drug names such as dalteparin, enoxaparin, heparin and tinzaparin, the suffix would indicate the nature of the word, and using suffix embeddings, we may be able to predict that the word refers to a heparin-like anticoagulant drug. To apply affix features to CEDI, the first $n$-characters (prefixes) and the last $n$-characters (suffixes) of each token are extracted. We excluded low-frequency prefix and suffix candidates from the set, since they would unlikely be true affixes. We assigned randomly initialized embeddings to each unique prefix and suffix. This embedding is concatenated with biLSTM output over characters of the token and pre-trained token embeddings for the token, and fed into the next biLSTM layer. Yadav et al. showed that deep affix features may improve overall performance of NER generally as well as in the medical domain such as SemEval 2013 task [3] and 2010 i2b2 clinical NER task [45] by utilizing three-character ($n = 3$) affixes with various thresholds for minimum frequency counts (e.g., 10, 20 or 50).

## 2.3 Attention mechanism

Attention mechanism directs the focus of the entity recognition system on a part of the input that is pertinent for the prediction task. The attention model was originally proposed by Bahdanau et al. [2] for machine translation and spread to other NLP tasks. It has been recently applied to NER tasks as well [33,49,55], but it has never been used for de-identification.

Using the attention mechanism, the biLSTM output (generated from the features such as character embeddings and token embeddings) is multiplied by the attention weights, yielding a context embedding for each token. The weights correspond to the amount of attention that the particular input "deserves" for the current prediction. This context embedding is concatenated (additive models) [26,34] or multiplied (multiplicative models) [46,51] with the hidden state of the corresponding timestamp and becomes probability vectors over labels $d_t$, yielding the final prediction for the token $y_t$.

Attention mechanisms may be utilized at multiple places in an NER system such as over character embeddings, over the token embeddings, and over context embeddings. An attention mechanism can be applied over character embeddings to find out which characters in the token are most critical to represent the token at the character level. If applied over token embeddings, it increases weights of informative tokens of the sentence for more accurate label prediction. CEDI uses the attention mechanism over concatenation of character, token and context embeddings.

## 2.4 Datasets and experiments

We test the de-identification performance of CEDI using datasets of 2006 i2b2 de-identification challenge (2006 shared task, henceforth) [43], 2014 i2b2/UTHealth shared task (2014 shared task, henceforth) [37] and 2016 CEGS N-GRID shared task (2016 shared task, henceforth) [36]. The 2006 shared task dataset contains 889 medical discharge summaries. The 2014 shared task dataset consists of 1,304 diabetic patient records. The 2016 shared task dataset consists of 1,000 psychiatric intake records. Table 1 summarizes the statistics of each dataset. As shown, 2014 dataset is 39% bigger than the 2006 dataset, which is the smallest. The 2016 dataset is twice as big as the 2014 dataset. The average sentence lengths of the 2006 and 2016 datasets are 54% and 42% longer than that of the 2014 dataset, respectively.

**Table 1: Overview of the de-identification datasets**

|  | 2006 Dataset | 2014 Dataset | 2016 Dataset |
|---|---|---|---|
| Records | 889 | 1,304 | 1,000 |
| Tokens | 580,800 | 805,118 | 1,862,452 |
| Tokens per record | 653 | 617 | 1,862 |
| Tokens per sentence | 11.25 | 7.3 | 10.4 |
| Sentences | 51,583 | 110,434 | 179,593 |
| Sentences per record | 58 | 85 | 180 |
| i2b2-PIIs | 19,669 | 28,872 | 34,364 |
| i2b2-PIIs per record | 22 | 22 | 34 |

The 2006 dataset contains 8 categories of PIIs, whereas the 2014 and 2016 datasets contain the same 28 types of PIIs in 7 categories as defined by the organizers of the i2b2 shared tasks (i2b2-PIIs, henceforth). The i2b2 PIIs are more extensive than the 18 types of HIPAA PIIs for better privacy protection. Originally, 2006 shared task provides 75% of the entire dataset as a training set. 2014 and 2016 shared task provide 60% of the datasets as training sets. The remaining 25% and 40% were used as test sets.

We split the original training set into training and validation sets using 2:1 ratio, respectively. We trained CEDI on the training set while checking its performance and tuning its parameters on the validation set. After reaching to a state where more training would not improve the performance of the system further, we tested the trained system on the test set and evaluated our approach based on the test results.

## 2.5 System structure

We tested the effectiveness of CEDI by integrating each component into the state-of-the-art de-identification system, NeuroNER, one by one. To exclude potential factors that may inadvertently affect the system performance, we used the original NeuroNER code, maintaining all their parameters.

Originally, NeuroNER uses character embeddings and pre-trained token embeddings as features. Token embeddings can be trained on general corpus such as Google News and Wikipedia as well as a domain specific corpus such as PubMed. While domain specific resources contain useful knowledge closely related to the relevant task, general resources may contain more generic semantics of English [47], therefore, the training resource should be carefully selected. NeuroNER achieved its highest de-identification performance using GloVe comprising general token embeddings from English version of the Wikipedia 2014 and English Gigaword Fifth Edition [30]. Therefore, CEDI incorporates both character embeddings and GloVe as well as its newly added $n$-grams based context embeddings.

Then we investigated the use of prefix and suffix features on the training data using 5-fold cross-validation. We found that prefix features increased the overall de-identification performance, but suffixes did not. Therefore, we only included the prefix features in our final system.

We also examined the benefits of an attention mechanism on the de-identification task. We tested it over the three noted locations, specifically, over characters, $n$-grams, and concatenated embeddings consisting of character embeddings, token embeddings, deep prefix features, and context embeddings. We did not use the attention mechanism over the components of the concatenated embeddings because the concatenated embeddings represent tokens better than their individual components. We found that only the attention over the concatenated embeddings improved the de-identification performance; therefore, we only used the attention mechanism applied to the concatenated embeddings.

## 2.6 Training and hyperparameters

We tuned hyperparameters on the training set using 5-fold cross-validation. Our selection for the final hyperparameters was as following: $n$-gram size = 10; character embeddings dimension = 25; prefix embeddings dimension = 25; prefix frequency threshold = 20; token embeddings dimension = 100; hidden layer dimension for LSTM over character = 25; hidden layer dimension for LSTM over prefix = 25; hidden layer dimension for LSTM over $n$-gram = 256; hidden layer dimension for LSTM over concatenation of {character, token, prefix and context embeddings} = 100; attention layer dimension = 50; dropout = 0.5; optimizer = SGD; learning rate = 0.02; maximum number of epochs = 100; early stopping = 15.

## 2.7 Evaluation metrics

We use the official evaluation scripts from the shared tasks. The primary metrics of the tasks are micro-averaged precision ($P$), recall ($R$) and F1-score ($F_1$). The organizers of the shared tasks applied the metrics in two different ways, entity-based and token-based. The entity-based evaluation considers a true positive count, only when the system successfully de-identifies the entire entity. In contrast, the token-based evaluation gives partial credit when a system de-identifies at least one token in the entity. It is debatable which evaluation method is more informative for de-identification purposes; however, we select entity-based evaluation, because it is stricter and commonly used for general NER tasks. Due to non-deterministic characteristic of the neural networks training, the performance from our suggested system can be different each time. We therefore repeat each experiment five times and report results that reflect the arithmetic mean of five runs per experiment. The statistical significance of F1-score differences from the baseline model is calculated using the approximate randomization test [28] using 9,999 shuffles.

## 3 Results

The results of CEDI experiments are summarized in Table 2. We use the original NeuroNER as our baseline and test the effect of adding each of the new components. As previously mentioned, NeuroNER uses character embeddings and token embeddings as features. The F1-scores of NeuroNER are 96.0, 91.7 and 88.1 for the 2006, 2014, and 2016 datasets, respectively. CEDI accomplishes F1-scores of 96.4 (+0.4), 92.2 (+0.7), and 89.5 (+1.4) on the 2006, 2014, and 2016 datasets, respectively. Adding deep prefix features on to CEDI (CEDI + $D_P$) further increases F1-score to 92.8 (+0.4) and 89.7 (+0.2) for the 2014 and 2016 datasets. Adding attention mechanism to this (CEDI + $D_P$ + $Att$) further increases the performance to 96.5 (+0.1) and 92.8 (+0.1) for the 2006 and 2016 datasets. All F1-score differences from the baseline are statistically significant at $\alpha$ level of 0.01.

**Table 2: Results of the experiments on the shared tasks test sets per feature set**

| Feature Sets | 2006 Dataset | | | 2014 Dataset | | | 2016 Dataset | | |
|---|---|---|---|---|---|---|---|---|---|
| | P (%) | R (%) | $F_1$ | P (%) | R (%) | $F_1$ | P (%) | R (%) | $F_1$ |
| NeuroNER (baseline) | 97.0 | 95.1 | 96.0 | 92.4 | 91.0 | 91.7 | 89.2 | 87.0 | 88.1 |
| CEDI | 97.7 | 95.1 | 96.4 | 93.4 | 91.5 | 92.4 | 91.1 | 88.0 | 89.6 |
| CEDI + $D_P$ | 97.2 | 95.6 | 96.4 | 94.5 | 91.1 | 92.8 | 91.4 | 88.0 | 89.7 |
| CEDI + $Att$ | 97.7 | 95.3 | 96.5 | 93.3 | 91.7 | 92.5 | 91.1 | 88.1 | 89.6 |
| CEDI + $D_P$ + $Att$ | 97.5 | 95.5 | 96.5 | 94.2 | 91.5 | 92.8 | 91.2 | 88.2 | 89.8 |

## 4 Discussion
### 4.1 Error analysis

The addition of context embeddings improves de-identification performance over the baseline. This is particularly true, where sentence boundary detection is incorrect, or the sentence boundaries in the original documents are unclear. For example, in these datasets, IDNUM and USERNAME entities are often recorded in a different line from the main contents. In those cases, systems typically recognize IDNUM and USERNAME as single token sentences, for which character embeddings and random token embeddings do not provide enough information for the correct prediction, but context embeddings do. Another example is tabulated forms, which are common in electronic health records, but sentence boundary detectors perform poorly on tabulated text data. They generally separate headings from contents and put them into multiple sentences resulting in many single-word sentences and losing all context information. This is particularly problematic with numeric data, where sentences may contain only text such as '##/##'. In these cases, de-identification systems without context embeddings are prone to misclassify the numeric token as a DATE because it does not have any information to differentiate the numerical measurement from a DATE mention. Because CEDI takes into account adjacent tokens in the previous and following sentences, it is able to find clues that lead to better predictions.

PIIs such as personal names and addresses are difficult to be recognized by symbolic AI systems that are not trained on the data. RNNs in general and CEDI in particular recognize such entities using the context around them. CEDI uses $n$-grams, where $n$ specifies the width of the context that can be adjusted according to the problem domain and the task. Performance comparison of our system against the baseline on such PII types is shown in Table 4. CEDI improved the baseline performance on all PII types except STREET. We suspect this was because of the small number of instances of this entity type, which was inadequate for training CEDI. The results show that the deep prefix feature, $D_P$, increases precision but not recall. This benefit manifests itself frequently more on capitalized words than all-lowercase words. Since prefix features filter out false positive predictions, high-frequency prefixes are less common in proper nouns (named entities) and this feature helps distinguish capitalized words that are not named entities.

The performance due to the attention mechanism depends on the length of the input sentence. Attention mechanism performs particularly well on longer sentences, in which the other embedding systems do not perform as well. Its highest performance improvement was on entity type HOSPITAL, which appears in relatively long and complete sentences.

**Table 4: Performance comparison with the baseline per difficult entity type**

| Entity type | NeuroNER | | | CEDI | | |
|---|---|---|---|---|---|---|
| | $P$ (%) | $R$ (%) | $F_1$ | $P$ (%) | $R$ (%) | $F_1$ |
| Patient | 86.5 | 83.8 | 85.2 | 87.6 | 83.2 | 85.3 |
| Doctor | 93.6 | 93.2 | 93.4 | 94.8 | 95.5 | 95.2 |
| Hospital | 87.3 | 76.2 | 81.4 | 87.3 | 76.8 | 81.7 |
| State | 88.3 | 89.4 | 88.8 | 89.4 | 91.5 | 90.4 |
| City | 83.8 | 87.3 | 85.5 | 83.0 | 89.0 | 85.9 |
| Street | 77.8 | 58.3 | 66.7 | 69.0 | 58.8 | 63.5 |

## 4.2 Comparison with ELMo

We compared CEDI's $n$-gram context embeddings to ELMo. The results are summarized in Table 3. CEDI generates an F1-score of 96.4 (+0.4), 92.4 (+0.7), and 89.6 (+1.5) on the 2006, 2014, and 2016 datasets, respectively. Whereas ELMo-incorporated-NeuroNER (NeuroNER + ELMo) generates an F1-score of 96.3 (+0.3), 92.8 (+1.1) and 89.0 (+0.9) on the 2006, 2014, and 2016 datasets. While ELMo achieves a slightly higher performance gain (+0.4) on the 2014 dataset based on F1-scores, CEDI performed better (+0.1 and +0.6 on the 2006 and 2016 dataset, respectively), using significantly less effort, time, and computational resources.

We also experimented with integrating ELMo into CEDI (CEDI + ELMo). This achieved the highest F1-score at 96.7 (+0.7), 93.3 (+1.6) and 90.5 (+2.4) on the 2006, 2014, and 2016 datasets, respectively. This implies that $n$-gram context embeddings generate complementary knowledge to ELMo embeddings and that understanding both general context and task specific context are helpful to increase the de-identification performance of the NER system.

**Table 3: Performance comparison between $n$-gram context embeddings vs ELMo**

| Feature Sets | 2006 Dataset | | | 2014 Dataset | | | 2016 Dataset | | |
|---|---|---|---|---|---|---|---|---|---|
| | $P$ (%) | $R$ (%) | $F_1$ | $P$ (%) | $R$ (%) | $F_1$ | $P$ (%) | $R$ (%) | $F_1$ |
| NeuroNER (baseline) | 97.0 | 95.1 | 96.0 | 92.4 | 91.0 | 91.7 | 89.2 | 87.0 | 88.1 |
| NeuroNER + ELMo | 96.9 | 95.7 | 96.3 | 94.3 | 91.4 | 92.8 | 90.9 | 87.2 | 89.0 |
| CEDI | 97.7 | 95.1 | 96.4 | 93.4 | 91.5 | 92.4 | 91.1 | 88.0 | 89.6 |
| CEDI + ELMo | 97.4 | 96.1 | 96.7 | 94.7 | 91.8 | 93.3 | 92.2 | 89.0 | 90.5 |

## 4.3 $n$-gram context embeddings on other tasks

The problems from erroneous sentence boundary detection and co-references across sentence boundaries are not limited to de-identification, but pertinent for all entity recognition tasks in the medical domain. The $n$-gram context embeddings which alleviate the problems may improve the system performance on other NER tasks. Therefore, we examined the usefulness of $n$-gram context embeddings on additional medical entity recognition tasks. We selected two datasets from previous shared tasks in the medical domain, the ShaRe/CLEF 2013 eHealth [38] Task 1 dataset (ShaRe/CLEF dataset, henceforth) and the fifth BioCreative challenge evaluation workshop [48] Track 3 CDR dataset (BC5CDR dataset, henceforth). The ShaRe/CLEF dataset contains disorder mentions from 299 de-identified clinical notes and BC5CDR contains disease mentions from 1,500 PubMed titles and abstracts. We set the training parameters identical to the ones we used for de-identification tasks but increased the size of the $n$-grams to 20 for the BC5CDR dataset because biomedical literature requires larger context than clinical notes. As shown in Table 4, the integration of $n$- gram context embeddings (CEDI) significantly improves entity recognition performance ($p < 0.01$) on both datasets. This implies $n$-gram context embeddings can be widely adapted as a feature for NER in the medical domain.

Table 4: Performance gains from $n$-gram context embeddings on other medical entity recognition tasks

| Feature Sets | ShaRe/CLEF | | | BC5CDR | | |
|---|---|---|---|---|---|---|
| | $P$ (%) | $R$ (%) | $F_1$ | $P$ (%) | $R$ (%) | $F_1$ |
| NeuroNER (baseline) | 78.1 | 70.3 | 74.0 | 84.4 | 80.6 | 82.4 |
| CEDI | 78.9 | 70.7 | 74.5 | 83.8 | 82.0 | 82.9 |

## 4.4 Training and inference time

One of the main challenges of using neural network de-identification systems is the time requirements. Training CEDI on i2b2 2014 training set with 790 records on a GPU with 2,304 cuda cores took 15 hours 13 mins on average, compared to 8 hours 2 mins for the baseline. Inference time per record was 0.5 seconds for CEDI and 0.2 seconds for NeuroNER.

## 5 Conclusions

We proposed a novel approach, a context-enhanced de-identification (CEDI) system, to mitigate the problems of modern NER systems that are based on biLSTM with CRF sequence optimizers (biLSTM-CRF). The performance of biLSTM-CRF NER systems depend on the accuracy of sentence boundary detection and they cannot capture dependencies beyond sentence boundaries. CEDI does not require sentence boundary detection and, using $n$-gram context embeddings, it can capture dependencies across sentence boundaries. CEDI also addresses the problems of OoV and ambiguous terms by introducing internally derived $n$-gram context embeddings. We demonstrated the benefits of CEDI using 2006 i2b2 de-identification challenge, 2014 i2b2/UTHealth shared task, and 2016 CEGS N-GRID shared task datasets. CEDI increased de-identification performance of NeuroNER on all three de-identification datasets. The performance gains from using $n$-gram context embeddings were even higher than that from utilizing ELMo on 2006 and 2016 dataset with significantly less computing resources. Using prefix feature and attention mechanisms further increased the de-identification performance of CEDI. Results of experiments with additional medical entity recognition datasets indicated that $n$-gram context embeddings can be widely adapted as a feature for many entity recognition tasks in medical domain. In this study, our focus was improving NeuroNER's performance with a new set of features of CEDI. Further studies are needed to test CEDI performance against recently introduced methods such as [1,8] which do not require sentence boundary detections, either. We plan to study and compare $n$-gram context embeddings with other contextualized embeddings methods such as [1,9].

**ACKNOWLEDGMENTS**

This work was supported by the Intramural Research Program of the National Institutes of Health, National Library of Medicine and by an appointment to the Science Education Programs at the National Institutes of Health, administered by ORAU through the U.S. Department of Energy Oak Ridge Institute for Science and Education.